%% file: main.tex
\documentclass[10pt,twocolumn,letterpaper]{article}
\usepackage{multirow}
\usepackage{booktabs}
\usepackage{xcolor}

\usepackage[pagenumbers]{wacv} 

\input{preamble}

\definecolor{wacvblue}{rgb}{0.21,0.49,0.74}
\usepackage[pagebackref,breaklinks,colorlinks,allcolors=wacvblue]{hyperref}

\def\wacvPaperID{814} 
\def\confName{WACV}
\def\confYear{2027}

\title{One Model, Two Worlds: Bidirectional Sonar–Optical Translation}

\author{
Shengji Jin\textsuperscript{1} \quad
Trung Tien Dong\textsuperscript{1} \quad
Ahmed Lamidi\textsuperscript{1} \quad
Chen Chen\textsuperscript{2} \quad
Xiaomin Lin\textsuperscript{1} \quad
Yi Sheng\textsuperscript{1}\\[4pt]
\textsuperscript{1}University of South Florida \quad
\textsuperscript{2}University of Central Florida \\
{\tt\small \{jins, dongt, ahmedlamidi, xlin2, sheng1\}@usf.edu} \quad
{\tt\small chen.chen@ucf.edu}
}

\begin{document}
\maketitle
\input{sec/0_abstract}

\input{sec/1_intro}
\input{sec/2_related_works}
\input{sec/3_methods}

\input{sec/4_experiments}

\input{sec/5_Conclusion}
\newpage
{
    \small
    \bibliographystyle{ieeenat_fullname}
    \bibliography{main}
}

\end{document}

%% file: preamble.tex
\makeatletter
\renewcommand\paragraph{\@startsection{paragraph}{4}{\z@}%
  {1ex \@plus .3ex \@minus .2ex}
  {-0.6em}
  {\normalfont\normalsize\bfseries}}
\makeatother

\usepackage{algorithm,algpseudocode}

\usepackage{microtype}
\makeatletter
\def\wacvsection{\@startsection{section}{1}{\z@}
   {-7pt plus -2pt minus -2pt}{4pt}{\large\bf}}
\def\wacvsubsection{\@startsection{subsection}{2}{\z@}
   {-6pt plus -2pt minus -2pt}{3pt}{\elvbf}}
\makeatother

%% file: sec/0_abstract.tex
\vspace{-0.5cm}
\begin{abstract}
Translating between imaging sonar and optical cameras is valuable for underwater perception, but supporting both directions with separate models duplicates storage and computation. A unified bidirectional model is therefore attractive, yet existing approaches largely treat the two directions symmetrically despite their fundamentally different image-formation physics. We argue that sharing a generative model does not require sharing the physics. We introduce the \textbf{Direction-Asymmetric Realism Bridge (DARB)}, which retains a shared diffusion-bridge trunk while routing direction-specific physical priors through asymmetric pathways: range-aware modulation for sonar-to-optical translation and polar ray-dependent processing for optical-to-sonar translation. We further show that symmetry in training is also costly: applying a common realism schedule reduces sonar-to-optical PSNR by $2.60$\,dB. Our \textbf{Adaptive Realism Supervision (ARS)} instead determines when, where, and how strongly perceptual supervision is applied from reconstruction quality and gradient balance. Together, DARB and ARS enable one bidirectional model to match the sonar-to-optical specialist within $0.11$\,dB PSNR, outperform the optical-to-sonar specialist by $0.70$ FID, and surpass two independently trained BBDMs on seven of eight metrics.

\vspace*{-0.5cm}
\end{abstract}

%% file: sec/1_intro.tex
\section{Introduction}
\label{sec:intro}


Underwater platforms routinely pair imaging sonar with optical cameras because the two sensing modalities compensate for each other's limitations \cite{kim2022underwater, ferreira2016underwater}.
Imaging sonar remains effective in turbid or poorly illuminated water, but its speckled, polar-geometry returns are difficult for human operators to interpret; optical cameras, in contrast, provide intuitive visual content but deteriorate rapidly under scattering and attenuation \cite{akkaynak2018revised,akkaynak2019sea}. This complementarity makes cross-modal translation valuable in both directions. Translating sonar into optical-like imagery can facilitate human supervision and interpretation, whereas translating optical imagery into sonar can generate additional acoustic training data, for which annotated corpora remain scarce \cite{aubard2025sonar}. Supporting both capabilities onboard, however, with two independent translation models would duplicate computation and storage. A unified bidirectional model therefore offers a more practical solution for embedded underwater platforms, where payload, memory, and compute are tightly constrained.

\begin{figure}[t]
\vspace{-0.5cm}
    \centering
    \includegraphics[width=\linewidth]{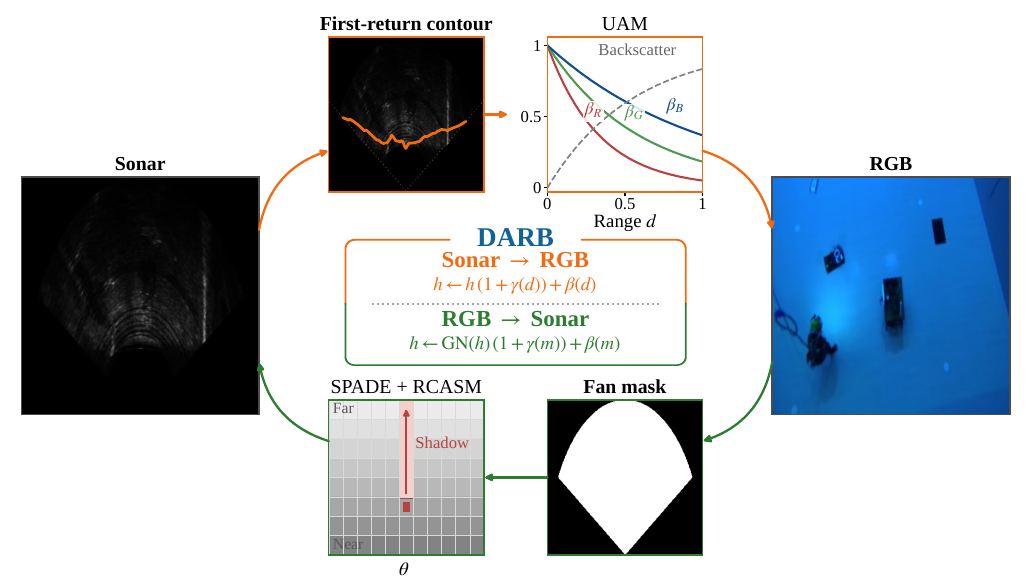}
    \caption{\textbf{Overview of DARB.} A single diffusion bridge serves both directions, each conditioned on a prior read out of its own source image. Sonar$\to$RGB (orange): the first-return contour gives a per-beam range field, applied by UAM. RGB$\to$Sonar (green): the fan-shaped support mask drives SPADE, and RCASM propagates occlusion along each bearing in polar coordinates.}
    \label{fig:overview}
    \vspace*{-0.3cm}
\end{figure}

Such a translator requires a framework in which neither direction is privileged. Conditional formulations privilege one: the source conditions a process synthesising the target, so the roles are not exchangeable. Diffusion bridges instead anchor the process at both endpoints, the natural substrate for a unified model~\cite{li2023bbdm,pmlr-v202-liu23ai,zhou2024denoising}. Existing unified bridges, however, remain data-driven: the two directions share every parameter and differ by a discrete label, so what separates them is left to be inferred from samples. We keep the shared substrate and put back only what the physics requires.

Sonar and optical images arise from fundamentally different image-formation processes and encode different physical properties of the same scene. Sonar observations are shaped by acoustic propagation, beam geometry, speckle, and shadowing, whereas optical appearance is governed by light transport, reflection, scattering, and attenuation. The two directions must therefore recover different missing information: sonar-to-optical translation must infer photometric appearance and fine texture that are not acoustically observed, while optical-to-sonar translation must synthesize acoustic responses and spatial structures that RGB imagery does not specify. We introduce the \textbf{Direction-Asymmetric Realism Bridge (DARB)}, which shares one diffusion-bridge trunk and makes asymmetric only what the imaging physics forces to be asymmetric. Spatially, this means routing a direction-specific physical prior into the decoder of each pathway (Fig.~\ref{fig:overview}). The optical path reads a per-beam range field that the sonar already carries and modulates features by it pointwise, as radiance decays and backscatter accumulates with range; the acoustic path reads the fan-shaped support of valid returns and the ray geometry along which a strong return shadows everything behind it, a relation that is local only in polar coordinates. A pointwise scaling and a path-dependent scan are not two settings of one operator, which is why the routing is direction-specific and the trunk is not.


Temporally, the same principle applies to the auxiliary perceptual objective that recovers realism, which should not act on both directions in the same way or from the same moment. Applying it uniformly from the start of training interferes with reconstruction before the bridge has learned a stable cross-modal mapping, and we observe that doing so lowers sonar-to-optical PSNR by $2.60$\,dB. The critical issue is therefore not the formulation of the realism objective but its schedule, which has to answer three questions: when the objective starts, which samples it acts on, and how strongly.

Exhaustively tuning these three coupled scheduling parameters would require a costly three-dimensional search and undermine the practical appeal of a unified framework. We therefore introduce \textbf{Adaptive Realism Supervision (ARS)}, which derives the supervision schedule from the evolving state of the model. A ratio between the gradient norms of the reconstruction and perceptual terms, read separately for each direction, answers the first and third: the objective is admitted as soon as reconstruction is the larger of the two forces, or after a fixed grace period if that does not happen, and is then weighted in proportion to that ratio. The second is answered per sample, realism being asked only of those the bridge already reconstructs well. Which of the two directions has to wait is decided by that measurement rather than by us. In summary, our contributions are:

\begin{itemize}
  \setlength{\itemsep}{0pt}
    \item \textbf{Direction-Asymmetric Routing:}
  We introduce DARB, a unified diffusion bridge for bidirectional sonar--optical translation that shares one trunk while routing a direction-specific physical prior into each pathway, capturing both formation processes without two independent translators.
  \item \textbf{Adaptive Realism Supervision (ARS):}
  ARS replaces a three-dimensional search over \emph{when} realism supervision starts, \emph{which samples} it acts on and \emph{how strongly}, with quantities already measured during training: gradient balance sets when and how strongly, per-sample reconstruction quality sets which.
  \item \textbf{Specialist-Level Performance with One Model:}
  On DeeperSense, DARB matches the Sonar$\to$RGB specialist to within 0.11\,dB PSNR, surpasses the RGB$\to$Sonar specialist by 0.7 FID, and beats two independently trained bridges on seven of eight metrics, so unified deployment need not cost translation quality.
\end{itemize}

%% file: sec/2_related_works.tex
\section{Related Work}
\label{sec:related}
We review how the field converged on one network, why sonar and optics break its assumption, and what follows per direction.
\subsection{The ``One Model'' Consensus}
\label{sec:rw_onemodel}

Serving both directions with a single network has been pursued across successive generations of translation models. Adversarial formulations solved paired and unpaired translation~\cite{isola2017image,zhu2017unpaired,huang2018multimodal,lin2022oysternet} and multi-domain generators consolidated several mappings into one network~\cite{choi2018stargan,choi2020stargan}, but a directional mapping fixes its input and output spaces, so both directions still require two generator pathways. Conditional diffusion~\cite{ho2020denoising,rombach2022high,lin2025odyssee} inherits the same asymmetry: the source steers a reverse process that synthesises the target~\cite{saharia2022palette,zhang2023adding}, and the two roles are not exchangeable. Diffusion bridges remove it by anchoring the stochastic process at both endpoints~\cite{li2023bbdm,pmlr-v202-liu23ai,zhou2024denoising}, which makes the intermediate trajectory symmetric~\cite{xue2025bibbdm}. Deterministic transports obtain the same property for free, since a learned ODE integrates in either direction~\cite{lipman2022flow,liu2022flow,tong2023improving,su2022dual,shi2023diffusion,kim2024unpaired}; stochastic bridges need an explicit mechanism, and three have appeared: BiBBDM supervises both endpoints with a six-channel objective~\cite{xue2025bibbdm}, BDBM masks one endpoint per sample and encodes direction by channel position~\cite{kieu2025bidirectional}, and CM-Diff pairs a domain-label embedding with modality-specific encoders~\cite{hu2025cm}. Measured by parameter efficiency, the unified bidirectional model is a settled problem.

While the capability to translate in both directions is established, the \emph{distance} between those domains remains largely unexamined. The domain pairs used to validate these unified models, sketches and shoes, semantic layouts and faces, infrared and visible light, are co-registered images formed by the same projection~\cite{xue2025bibbdm,kieu2025bidirectional,hu2025cm}. In such near-bijective settings the representations one direction needs largely serve the other, which is what makes one set of weights sufficient. Even at that short distance the two directions have not converged into a single problem: the infrared and visible literature has developed along separate lines for each direction, one pursuing structural detail and the other thermal realism~\cite{hu2025cm}. Asymmetric treatment is therefore already recognised at the level of methods, yet no unified model has been designed to internalise it within shared weights, across domains that do not even share an imaging equation.

\subsection{The ``Two Worlds'' Reality}
\label{sec:rw_twoworlds}

A forward-looking sonar does not photograph a scene; it times echoes. Each azimuthal beam records intensity integrated over a vertical fan of rays. Consequently, elevation is collapsed, and a single pixel in the resulting range-bearing image corresponds to an arc of three-dimensional points rather than a single spatial coordinate~\cite{aykin2016three,qadri2024aoneus}. Three distinct properties arise from this active sensing mechanism. First, returned amplitudes carry multiplicative speckle whose distribution is a property of the surface but whose realisation is random~\cite{huang2020speckle}. Second, occlusion appears as shadows extending outward along the ray from the transducer, governed by sensor geometry rather than the image plane~\cite{wang20232d}. Third, the echo carries no wavelength information, whereas optical appearance is dictated by the water column, where attenuation and backscatter are wavelength- and range-dependent, making colour a property of the medium as much as of the object~\cite{akkaynak2018revised,akkaynak2019sea}.

In terms of information content, these physical discrepancies fracture the bidirectional task into two fundamentally asymmetric problems. Because elevation is integrated and wavelength is absent, translating optical images to sonar represents a many-to-one mapping; the network must learn what to discard while synthesising a random speckle realisation unconstrained by the input. Conversely, translating sonar to optical imagery is a heavily under-determined one-to-many mapping, requiring the hallucination of photometric texture absent from the source. Existing work accordingly treats the two as separate tasks: sonar-to-optical synthesis for diver monitoring~\cite{wehbe2022sonar}, sonar synthesis for data augmentation~\cite{wang20232d,xu2026forward}, and surveys that analyse one direction at a time~\cite{aubard2025sonar,zhang2020quality}. Sonar and optical imagery share none of this, which is what makes them two worlds rather than two views.

\subsection{Navigating the Gap between ``Two Worlds''}
\label{sec:rw_navigating}

This physical asymmetry raises three challenges, and existing work approaches each from one side only: unified models treat the two directions as one problem, while physics-informed generators solve one direction at a time. The first is the spatial prior. Feature modulation injects structure without bottlenecking the input~\cite{park2019semantic,perez2018film}, and physics-informed generation goes further by building the formation model into the generator, for infrared synthesis~\cite{mao2026pid} and for ultrasound, where the diffusion process itself is modelled as wave propagation~\cite{dominguez2024diffusion}. The principle transfers to our setting; the priors do not, each being bound to a single target domain and unidirectional by construction. The second is auxiliary supervision. Perceptual and adversarial objectives recover textures that reconstruction losses smooth away~\cite{johnson2016perceptual,zhang2018unreasonable,blau2018perception}, and diffusion training weights them along the noise axis~\cite{choi2022perception,hang2023efficient} or anneals them over the schedule. Existing unified schedules apply one such rule to all samples regardless of direction.

The third is evaluation. The structural similarity index rests on local correlation~\cite{wang2004image}, so it collapses on content that is statistically accurate but spatially decorrelated, a known failure mode for resampled textures~\cite{ding2020image} and for coherent sensor speckle~\cite{zhang2020quality}. Bridging two physically incongruent worlds forces all three to be made per direction: which prior each requires (Sec.~\ref{sec:method_routing}), when each is supervised (Sec.~\ref{sec:method_ars}), and against which reference each is read (Sec.~\ref{sec:exp_instrument}).

%% file: sec/3_methods.tex
%
\section{Method}
\label{sec:method}

Let $(a,b)$ be a co-registered sonar and optical pair, encoded by a frozen autoencoder into latents of identical shape. We seek a single model that serves $a \mapsto b$ and $b \mapsto a$ at the cost of one model. Write $d \in \{0,1\}$ for the requested direction, $x_0$ for the target latent and $y$ for the source, so that $(x_0,y) = (b,a)$ when $d=0$ and $(a,b)$ when $d=1$.

The latent Brownian bridge we build on is symmetric in $d$ by construction (Sec.~\ref{sec:method_prelim}). DARB reintroduces asymmetry in exactly two places: the operators each direction is routed through (Sec.~\ref{sec:method_routing}) and the schedule under which a realism term is applied to it (Sec.~\ref{sec:method_ars}). Everything else is shared. Fig.~\ref{fig:framework} shows the architecture; the training procedure is given in the supplement.

\subsection{Preliminaries: the Latent Brownian Bridge}
\label{sec:method_prelim}

\begin{figure*}[t]
\vspace{-0.5cm}
    \centering
    \includegraphics[width=\textwidth]{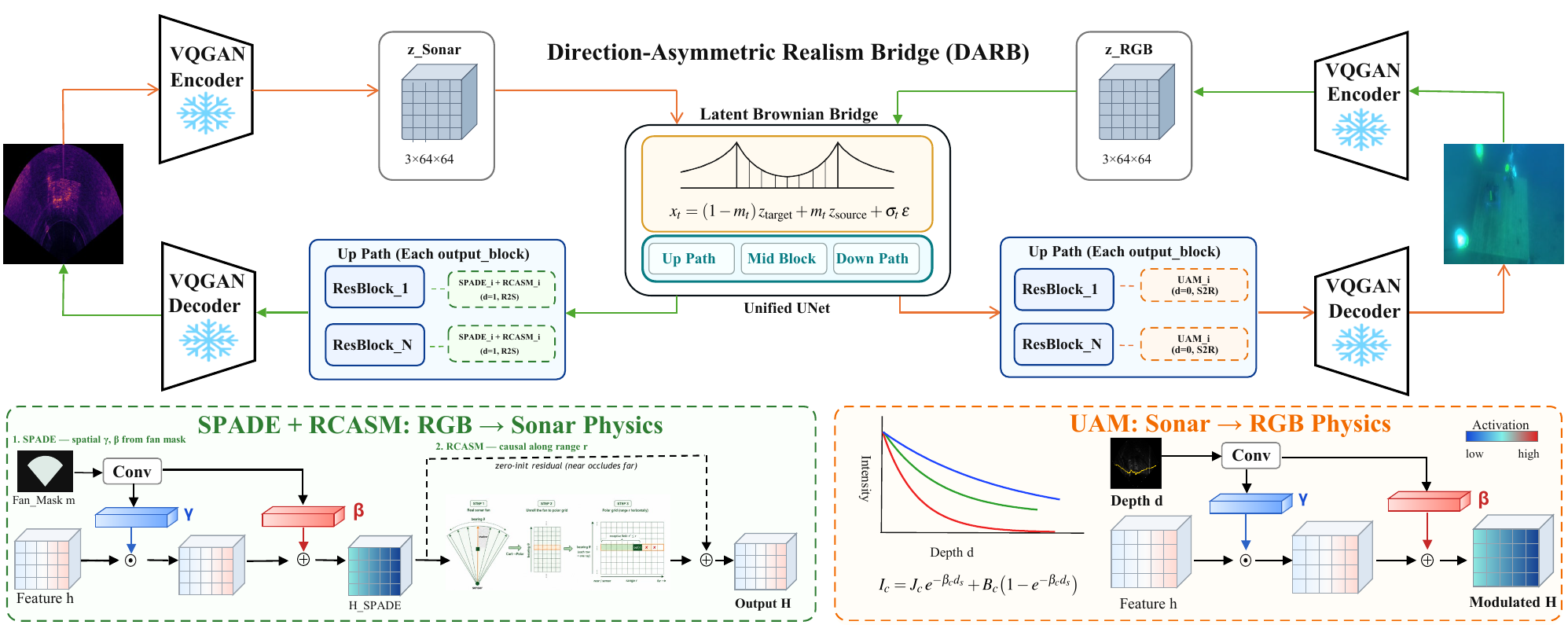}
    \caption{\textbf{DARB.} A single trunk serves both directions. The direction scalar $d$ enters through its own zero-initialised AdaGN branch and selects which physical operators the forward pass is routed through: UAM for the optical target, SPADE and RCASM for the acoustic one.}
    \label{fig:framework}
    \vspace*{-0.2cm}
\end{figure*}
In the shared latent space of a frozen autoencoder~\cite{esser2021taming,rombach2022high}, a Brownian bridge interpolates directly between the two latents rather than between data and pure noise~\cite{li2023bbdm}. With $m_t = t/T$ and $\epsilon \sim \mathcal{N}(0,\mathbf{I})$,
\begin{equation}
x_t = (1-m_t)\,x_0 + m_t\,y + \sigma_t\,\epsilon, \qquad \sigma_t^2 = 2\,m_t(1-m_t).
\label{eq:forward}
\end{equation}
so the state recovers $x_0$ at $t=0$ and $y$ at $t=T$. The model $\mathcal{O}_\theta$ predicts the displacement to the target, and $\lVert \cdot \rVert_1$ denotes a mean over elements throughout:
\begin{equation}
\mathcal{O}_t = m_t\,(y-x_0) + \sigma_t\,\epsilon, \qquad \mathcal{L}_{\mathrm{br}} = \bigl\| \mathcal{O}_t - \mathcal{O}_\theta(x_t, t, d) \bigr\|_1 .
\label{eq:objective}
\end{equation}
Subtracting the prediction from the state gives an explicit draft at any step:
\begin{equation}
\hat{x}_0 = x_t - \mathcal{O}_\theta(x_t, t, d), \qquad \hat{x}_0 - x_0 = \mathcal{O}_t - \mathcal{O}_\theta .
\label{eq:draft}
\end{equation}
The right-hand identity is exact and carries no $m_t$ factor, so the draft inherits the training error uniformly along the bridge. An $\epsilon$-parameterisation instead amplifies that error by $\sigma_t/(1-m_t)$, which reaches $44.7$ at the far endpoint. The displacement form therefore gives an $\hat{x}_0$ stable enough to attach an auxiliary objective to (Sec.~\ref{sec:method_ars}).

Eqs.~\eqref{eq:forward} to~\eqref{eq:draft} never reference $d$: the symmetry belongs to the stochastic process, not to the physical problem, so a unified model has to put the asymmetry back deliberately.

\subsection{Direction-Asymmetric Routing}
\label{sec:method_routing}

\paragraph{Isolating the direction signal.}
The conventional route folds the direction $d$ into the timestep embedding as a label term, where a binary signal shares one projection with a continuous, high-variance one. We keep that term but additionally give $d$ its own modulation stage in every residual block. After the timestep modulation, the feature map passes through a second, direction-only affine transform:
\begin{equation}
\begin{gathered}
h \leftarrow \bigl[\mathrm{GN}(h)\,(1+s_t) + b_t\bigr]\,(1+s_d) + b_d, \\
(s_d, b_d) = \mathcal{M}_d(e_d),
\end{gathered}
\label{eq:adagn}
\end{equation}
where $e_d$ is the direction embedding and $\mathcal{M}_d$ is zero-initialised, so training begins as an exactly symmetric bridge and departs from it only where the data demands. The branch costs $12.9$\,M parameters, the smallest of the asymmetries we introduce. One direction is drawn per batch rather than per sample, so a single read of $d$ can route the entire forward pass through the physical operators below.

\paragraph{Optical physics asks for value-domain modulation (UAM).}
Underwater optical appearance follows a formation model in which radiance decays multiplicatively with range and is compounded by additive backscatter~\cite{akkaynak2018revised,akkaynak2019sea}: $I_c = J_c e^{-\beta_c d_s} + B_c(1-e^{-\beta_c d_s})$, with $d_s$ the scene depth and $\beta_c$ a wavelength-dependent coefficient. The variable this model needs is one the sonar records: we extract $d_s$ from the sonar itself, as the range of the first return along each beam, precomputed per pair and normalised to $[0,1]$. In the sonar-to-optical direction the source is the sonar, so the prior draws on nothing beyond the input.

The Underwater Attenuation Module conditions the optical path on this field by mirroring the model's form, one multiplicative and one additive term per channel:
\begin{equation}
\mathrm{UAM}(h, d_s) = h \odot \bigl(1 + \gamma(d_s)\bigr) + \beta(d_s),
\label{eq:uam}
\end{equation}
where $\gamma$ and $\beta$ come from a lightweight convolutional stack and are zero-initialised. UAM applies this modulation without normalising $h$ first: attenuation acts on absolute intensity, and normalisation would discard the very scale the module exists to modulate. The depth field is resampled bilinearly, since it is a smooth scene signal rather than a boundary.

\paragraph{Acoustic physics asks for geometric routing (SPADE and RCASM).}
The acoustic path faces constraints of a different kind. The first is the fan-shaped support of valid returns, across the four captures of training data. We impose it with Spatially Adaptive Normalization (SPADE)~\cite{park2019semantic}, conditioned on the per-image binary support mask: the feature map is normalised without learned affine parameters and then rescaled and shifted by a zero-initialised stack driven by the mask. The mask is resampled with nearest neighbours, because bilinear interpolation would soften the one edge the constraint exists to draw.

The second is acoustic occlusion. A strong return at range $r$ and bearing $\theta$ shadows every larger range along that bearing: a prefix relation along a ray whose direction rotates with bearing. A translation-invariant convolution has one kernel orientation for the whole image, so this relation is not expressible in the Cartesian plane. The Ray-Casting Acoustic Shadow Module (RCASM) therefore changes coordinates. It warps the feature map $h$ and the support mask $m$ onto a canonical polar $(r,\theta)$ grid wide enough to cover every sensor in the data, applies a one-dimensional causal convolution along $r$, padded only on the near side so that no far-range information leaks backwards, and returns the result as a residual through a zero-initialised projection $\mathcal{P}$:
\begin{equation}
h \leftarrow h + \mathcal{W}^{-1}\Bigl(\mathcal{P}\bigl(\phi\bigl(\mathrm{Conv}^{\mathrm{causal}}_{k\times 1}\bigl[\mathcal{W}(h);\,\mathcal{W}(m)\bigr]\bigr)\bigr)\Bigr),
\label{eq:rcasm}
\end{equation}
with $\mathcal{W}$ the Cartesian-to-polar transform. The warped mask rides along so that the convolution knows where the true fan of each image lies within the canonical grid.

\paragraph{Why the two cannot be one operator.}
The two priors place contradictory demands, not merely different ones. UAM must not normalise, because attenuation lives in the absolute scale of the features, while SPADE must normalise first, because its modulation is defined on whitened features. Attenuation is also pointwise, whereas shadowing is path-dependent, conditioned on all nearer ranges along a bearing; a pointwise operator stays pointwise in any coordinates, which is why only the acoustic prior needs a change of them. Routing the forward pass through direction-specific operators while keeping the diffusion trunk shared is what lets one model satisfy both sets of demands at once.

\subsection{Adaptive Realism Supervision (ARS)}
\label{sec:method_ars}
The bridge loss fixes structure but smooths away the high-frequency speckle and texture that make a cross-modal sample look real. Perceptual objectives~\cite{johnson2016perceptual,zhang2018unreasonable} recover it, but one transplanted unchanged into a shared bidirectional trunk degrades both directions at once (Sec.~\ref{sec:exp_ablation}). What fails is not the objective but its schedule: when the term starts, which samples it acts on, and how strongly. ARS answers all three from quantities the training loop already produces.

Both readings are taken at $W_{\mathrm{out}}$, the last trainable weight of the output head, where the two objectives meet. Every $P$ steps we retain the training graph, take two local backward passes, and form the force ratio
\begin{equation}
s_d = \frac{\lVert \nabla_{W_{\mathrm{out}}} \mathcal{L}_{\mathrm{br}} \rVert}{\lVert \nabla_{W_{\mathrm{out}}} \mathcal{L}_{\mathrm{perc}} \rVert},
\label{eq:sratio}
\end{equation}
clipped to $[s_{\min}, s_{\max}]$ and held per direction as a moving average. The probe differentiates one weight matrix rather than running a second training step.

\paragraph{When it starts.}
For the first $E_{\mathrm{win}}$ epoch-equivalents of a direction the perceptual term is formed but not applied. As that window closes, the median of $s_d$ is compared against unity, the point at which the two forces are equal: a direction whose reconstruction gradient is the larger admits the term at once, and a direction whose perceptual gradient already dominates defers it to a fixed epoch $E_{\mathrm{def}}$. Unity is the only scale-free threshold for a ratio of two measured quantities. The measurement therefore decides which direction waits, not how long (Sec.~\ref{sec:exp_ablation}).

\paragraph{Which samples it acts on.}
A perceptual term is informative only on a draft that already carries a layout. Rather than legislate where such drafts sit on the diffusion axis, ARS reads them off the batch: the draft error $e_i = \lVert \hat{x}_0^{(i)} - x_0^{(i)} \rVert_1$ is the current training residual, so selection costs no extra forward or backward pass, and the $\kappa$ samples of smallest error receive the term:
\begin{equation}
\mathcal{K} = \operatorname*{arg\,min}_{\lvert \mathcal{K} \rvert = \kappa} \; \textstyle\sum_{i \in \mathcal{K}} e_i .
\label{eq:select}
\end{equation}
The alternative, a fixed cut $t \le t_c$, would settle that region before training and impose the same one on both directions.

\paragraph{How strongly it acts.}
The weight is set in proportion to the same ratio,
\begin{equation}
\lambda_d = \mathrm{clip}\bigl(k\,s_d,\; \lambda_{\min},\, \lambda_{\max}\bigr).
\label{eq:dose}
\end{equation}
The proportionality is what makes this more than a heuristic. Writing $\rho_d$ for the share of the update at $W_{\mathrm{out}}$ that the perceptual term contributes and substituting $\lambda_d = k s_d$,
\begin{equation}
\rho_d = \frac{\lambda_d \lVert \nabla \mathcal{L}_{\mathrm{perc}} \rVert}{\lVert \nabla \mathcal{L}_{\mathrm{br}} \rVert + \lambda_d \lVert \nabla \mathcal{L}_{\mathrm{perc}} \rVert} = \frac{k}{1+k},
\label{eq:rho}
\end{equation}
so one constant fixes the realism budget as a fraction of the reconstruction force, identically in both directions and independently of how  loss scales drift as the learning rate decays. A fixed $\lambda$ has neither property, holding one direction's balance at the other's cost and drifting over training.

\paragraph{The objective.}
With $\mathcal{D}$ the frozen decoder, $x^{\mathrm{pix}}$ the original image whose latent is $x_0$, $e$ the current epoch and $e_\star^{(d)}$ the onset of direction $d$,
\begin{equation}
\mathcal{L} = \mathcal{L}_{\mathrm{br}} + \lambda_d \, \mathbf{1}\!\left[e \ge e_\star^{(d)}\right] \frac{1}{\kappa} \sum_{i \in \mathcal{K}} \ell_{\mathrm{perc}}\!\bigl(\mathcal{D}(\hat{x}_0^{(i)}),\, x^{\mathrm{pix},(i)}\bigr).
\label{eq:total}
\end{equation}
The reference is the original image rather than its reconstruction $\mathcal{D}(x_0)$, so the decoder's own error stays inside the target instead of cancelling out. What remains fixed is the budget $k$, the deferred onset $E_{\mathrm{def}}$ and the clamp on $\lambda_d$; none is searched on the unified model, $k$ being back-solved once from the two single-direction optima and $E_{\mathrm{def}}$ acting only when the ratio declines to fire. Searching instead means a grid over onset, selection and dose in each direction separately; the supplement lists our values.

\begin{table*}[t]
\vspace{-0.5cm}
\centering
\small
\setlength{\tabcolsep}{4.4pt}
\begin{tabular}{l c c cccc cccc}
\toprule
\multirow{2}{*}{Method} & \multirow{2}{*}{\#Models} & \multirow{2}{*}{Ep./dir.} & \multicolumn{4}{c}{Sonar $\rightarrow$ RGB} & \multicolumn{4}{c}{RGB $\rightarrow$ Sonar} \\
\cmidrule(lr){4-7} \cmidrule(lr){8-11}
 & & & PSNR $\uparrow$ & SSIM $\uparrow$ & LPIPS $\downarrow$ & FID $\downarrow$ & PSNR $\uparrow$ & SSIM $\uparrow$ & LPIPS $\downarrow$ & FID $\downarrow$ \\
\midrule
\multicolumn{11}{l}{\emph{Two models, one trained per direction}} \\
pix2pix~\cite{isola2017image}      & 2 & 100 & 23.36 & 0.856 & 0.183 & 70.22  & 23.24 & 0.583 & 0.215 & 106.21 \\ 
MUNIT~\cite{huang2018multimodal}        & 2 & 100 & 22.39 & 0.848 & 0.281 & 126.49 & 18.28 & 0.397 & 0.491 & 160.33 \\ 
OT-CFM~\cite{tong2023improving}        & 2 & 100 & 30.14 & 0.929 & 0.071 & 29.68  & 31.14 & 0.735 & 0.061 & 28.98 \\ 
BBDM~\cite{li2023bbdm}        & 2 & 100 & 30.61 & 0.936 & 0.047 & 14.04 & 32.15 & 0.762 & 0.050 & 17.29 \\
Ours, direction-specific      & 2 & 100 & \underline{32.40} & \underline{0.943} & \underline{0.034} & \underline{10.87} & \underline{32.16} & \underline{0.773} & 0.045 & 15.21 \\
\midrule
\multicolumn{11}{l}{\emph{One model serving both directions}} \\
BDBM~\cite{kieu2025bidirectional}      & 1 & 100 & 13.83 & 0.795 & 0.274 & 55.69 & 17.22 & 0.307 & 0.168 & 42.29 \\
BiBBDM~\cite{xue2025bibbdm}   & 1 & 100 & 30.87 & 0.934 & 0.047 & 12.19 & 31.91 & 0.756 & 0.049 & 16.27 \\
BiBBDM                        & 1 & 200 & 31.77 & 0.937 & 0.040 & 11.66 & \textbf{32.14} & 0.763 & 0.046 & 14.98 \\
DARB (ours)                   & 1 & 50  & 30.12 & 0.926 & 0.044 & 12.87 & 31.39 & 0.767 & 0.048 & 15.56 \\
\textbf{DARB (ours)}          & 1 & 100 & \textbf{32.29} & \textbf{0.941} & \textbf{0.035} & \textbf{11.08} & 32.10 & \textbf{0.769} & \textbf{\underline{0.044}} & \textbf{\underline{14.51}} \\
\midrule
$\Delta$ vs.\ BBDM                    & & & $+1.68$ & $+0.005$ & $-0.012$ & $-2.96$ & $-0.05$ & $+0.007$ & $-0.006$ & $-2.78$ \\
$\Delta$ vs.\ BiBBDM at $2\times$     & & & $+0.52$ & $+0.004$ & $-0.005$ & $-0.58$ & $-0.04$ & $+0.006$ & $-0.002$ & $-0.47$ \\
$\Delta$ vs.\ direction-specific      & & & $-0.11$ & $-0.002$ & $+0.001$ & $+0.21$ & $-0.06$ & $-0.004$ & $-0.001$ & $-0.70$ \\
\bottomrule
\end{tabular}
\caption{\textbf{Bidirectional sonar and optical translation on DeeperSense}, 1181 held-out pairs. Methods reported in one direction only are in the supplement. Ep./dir.\ is the budget each direction receives, so an alternating model is listed at half its raw epoch count. \textbf{Bold} marks the best single-model result and \underline{underline} the best result overall.}
\label{tab:main}
\vspace{-0.2cm}
\end{table*}

\subsection{Training and Inference}
\label{sec:method_fixed}




Sampling follows the standard bridge reverse process with $d$ held fixed, so a single trained model serves either direction.

To keep the asymmetry bounded, everything not named in Sec.~\ref{sec:method_routing} is shared: the trunk weights, the bridge schedule, the sampler and its step count, and every constant of ARS, including the budget $k$, the threshold, the window and the deferred onset. Two quantities do differ between directions, the onset $e_\star^{(d)}$ and the weight $\lambda_d$, but neither is specified. Both are what one rule returns when applied to each direction separately, and a direction whose force ratio behaved like the other's would receive the same schedule. We do not construct asymmetry by hand; we route the priors the physics requires and let the schedule follow the measurement. Sharing a trunk does not make the two directions interfere, as Sec.~\ref{sec:exp_main} shows; it leaves less spare capacity per direction, which is why realism must be metered, not applied outright.


%% file: sec/4_experiments.tex
\section{Experiments}
We ask how one model compares with specialists and other bridges, which of its parts matters, and how far the comparison carries.

\paragraph{Data and baselines.}

We use DeeperSense~\cite{wehbe2022sonar}, which provides co-registered forward-looking sonar and optical pairs across four captures of differing native resolution. We resize all images to $256 \times 256$, hold out 1181 pairs and train on the remaining 9448. Baselines span adversarial translation (Pix2Pix~\cite{isola2017image}, MUNIT~\cite{huang2018multimodal}, UNSB~\cite{kim2024unpaired}), conditional generation (ControlNet~\cite{zhang2023adding}, Palette~\cite{saharia2022palette}, OT-CFM~\cite{tong2023improving}) and diffusion bridges (BBDM~\cite{li2023bbdm}, BDBM~\cite{kieu2025bidirectional}, BiBBDM~\cite{xue2025bibbdm}); all but BDBM and BiBBDM need one model per direction, and UNSB, ControlNet and Palette are reported for one direction only, in the supplement. Every bridge uses the same autoencoder, schedule and sampler as DARB, so the comparison isolates how direction is expressed. Each baseline runs under its released configuration; for BDBM this means the noise-prediction objective.
 
\paragraph{Training details.}
DARB uses a frozen VQGAN~\cite{esser2021taming} with a downsampling factor of four, 1000 diffusion steps in training and 200 sampling steps at inference, and requires 296.94\,M trainable parameters, a $37\%$ reduction relative to two BBDM bridges at 237.10\,M each. We train with Adam at 1e-4 and an effective batch size of 32 on a single RTX 5090. Every model receives 29{,}500 gradient updates per direction. Each configuration is trained once under a single seed; remaining hyperparameters are in the supplement.
 
\paragraph{Evaluation.}
We report PSNR, SSIM~\cite{wang2004image}, LPIPS~\cite{zhang2018unreasonable} and FID~\cite{heusel2017gans} on decoded RGB images. LPIPS is also an auxiliary training objective in DARB, so FID is the only perceptual measure no model was trained against and we read it as primary. All four are reported per direction and never averaged, because the attainable range of a metric can differ between the domains; Sec.~\ref{sec:exp_instrument} measures that range for SSIM.

\begin{figure*}[t]
\vspace{-0.2cm}
  \centering
  \includegraphics[width=0.9\textwidth]{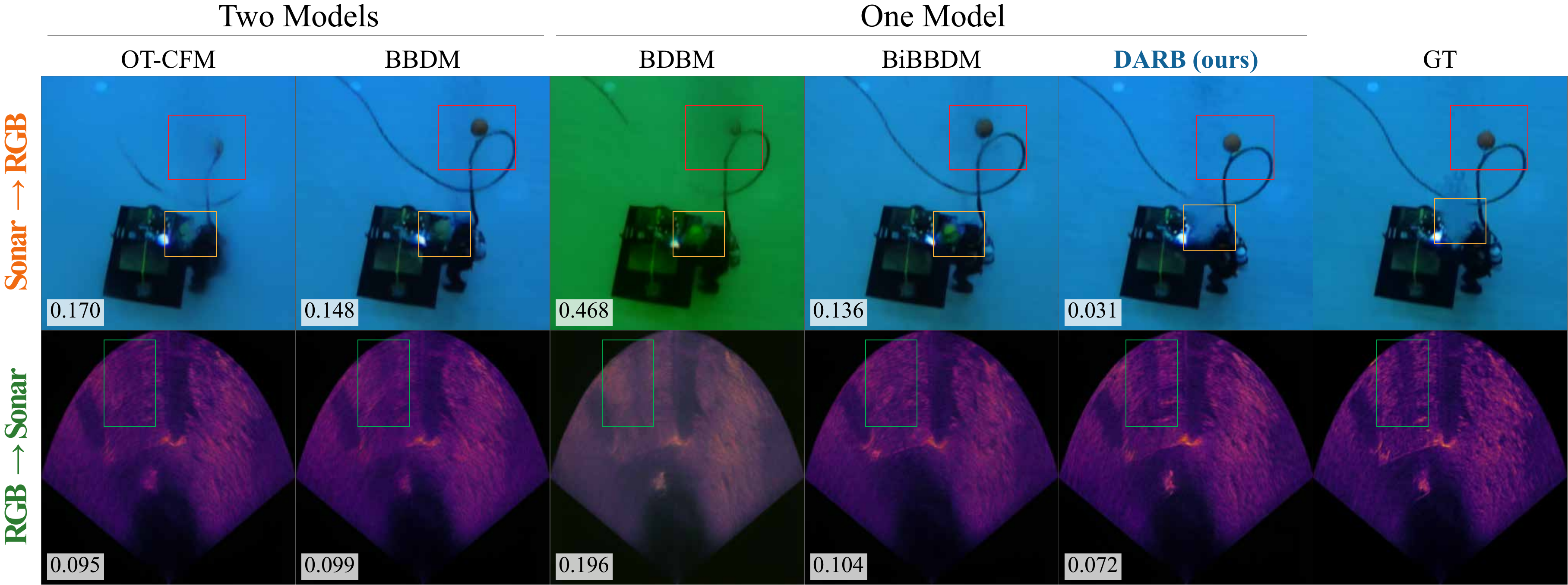}
  \caption{\textbf{Both directions on one held-out pair.} Column groups separate methods that need one trained model per direction from those that serve both with one. Per-image LPIPS is inset. Sonar renderings share a colour map and scale.}
  \label{fig:qual}
  \vspace{-0.2cm}
\end{figure*}

\subsection{Quantitative results.}\label{sec:exp_main}
Tab.~\ref{tab:main} asks what one model gives up against progressively stronger alternatives. The adversarial and flow-based baselines are furthest behind: OT-CFM, the strongest, trails DARB by $2.15$\,dB PSNR and $18.60$ FID in the sonar-to-optical direction, because they synthesise the target from noise or a compressed code, whereas a bridge starts at the source image. Against BBDM, the strongest per-direction bridge, DARB gains $1.68$\,dB PSNR and $2.96$ FID forward and $2.78$ FID in reverse, and trails only by $0.05$\,dB in reverse PSNR. Serving both directions from one trunk therefore costs one cell in eight, at $37\%$ fewer parameters.

Among single-model methods, DARB leads BiBBDM in all eight cells at a matched budget and in seven of eight when BiBBDM trains twice as long, the exception being reverse PSNR at $0.04$\,dB. BiBBDM lets direction enter only through which endpoint is supervised, while DARB reserves $12.9$\,M for a direction-specific pathway. BDBM reaches only $13.83$\,dB and $55.69$ FID forward under its released noise-prediction objective, which Sec.~\ref{sec:method_prelim} identifies as ill-conditioned as $t$ approaches $T$. The strictest comparison is two direction-specific models with the same routing and supervision, which bound what a shared trunk can recover: DARB stays within $0.11$\,dB PSNR and $0.002$ SSIM of them forward and is ahead on both perceptual metrics in reverse, at $0.044$ against $0.045$ LPIPS and $14.51$ against $15.21$ FID. Halving the per-direction budget costs $2.17$\,dB, so the match does not follow from an arbitrary stopping point.

Per-image metrics do not say whether the acoustic intensities are right, so we pool the generated sonar over the held-out set (Fig.~\ref{fig:hist}). At full range only BDBM separates, placing $39\%$ of its pixels between intensities $50$ and $120$ where the reference has $4\%$. The ordering is read above intensity $9$, below which every latent method displaces the saturated background by one intensity level: DARB's total variation distance to the reference is $0.024$, against $0.042$ for the next closest bridge and $0.34$ for BDBM, unchanged for any floor between $3$ and $25$. The acoustic direction is therefore distributionally close, not merely structurally plausible.
 
\subsection{Qualitative results.}
Fig.~\ref{fig:qual} shows both directions on one held-out pair. BDBM misses the colour of both domains at once, casting the optical scene green and filling the region outside the fan with grey, the signature of one output distribution covering two domains. The float (red) is absent from OT-CFM and BDBM, displaced in BBDM and BiBBDM; only DARB places it where the reference does. The diver's head (orange) is occluded by bubbles, and only DARB recovers it. In the acoustic direction the shadow behind a strong return (green) appears only in DARB, the geometric propagation RCASM is built for. DARB carries the lowest per-image LPIPS in both directions, $0.031$ and $0.072$.



\subsection{Ablation}
\label{sec:exp_ablation}

\begin{table*}[t]
\centering
\small
\begin{minipage}[t]{0.485\textwidth}
\centering
\setlength{\tabcolsep}{3.2pt}
\begin{tabular}{l cccc}
\toprule
\multicolumn{5}{c}{\textbf{Sonar $\rightarrow$ RGB}} \\
\midrule
Configuration & PSNR $\uparrow$ & SSIM $\uparrow$ & LPIPS $\downarrow$ & FID $\downarrow$ \\
\midrule
\multicolumn{5}{l}{\footnotesize\emph{Dedicated model}} \\
Bridge only                        & 30.61 & 0.936 & 0.047 & 14.04 \\
$+$ UAM prior                      & 32.45 & 0.946 & 0.042 & 18.47 \\
\;$\hookrightarrow$ realism, always on      & 31.48 & 0.938 & 0.036 & 11.13 \\
\;$\hookrightarrow$ realism, gated in $t$   & 31.01 & 0.935 & 0.038 & 11.70 \\
\;$\hookrightarrow$ realism, gated in epoch & 32.40 & 0.943 & 0.034 & 10.87 \\
\midrule
\multicolumn{5}{l}{\footnotesize\emph{Shared model, both priors on, cumulative}} \\
Shared trunk, joint AdaGN          & 31.08 & 0.936 & 0.044 & 14.20 \\
$+$ separate AdaGN                 & 31.74 & 0.939 & 0.041 & 14.33 \\
$+$ realism, always on             & 29.14 & 0.924 & 0.057 & 14.15 \\
$+$ realism, gated in $t$           & 31.38 & 0.938 & 0.038 & 11.73 \\
\textbf{$+$ realism, gated in epoch} & 32.29 & 0.941 & 0.035 & 11.08 \\
\bottomrule
\end{tabular}
\end{minipage}
\hfill
\begin{minipage}[t]{0.485\textwidth}
\centering
\setlength{\tabcolsep}{3.2pt}
\begin{tabular}{l cccc}
\toprule
\multicolumn{5}{c}{\textbf{RGB $\rightarrow$ Sonar}} \\
\midrule
Configuration & PSNR $\uparrow$ & SSIM $\uparrow$ & LPIPS $\downarrow$ & FID $\downarrow$ \\
\midrule
\multicolumn{5}{l}{\footnotesize\emph{Dedicated model}} \\
Bridge only                        & 32.15 & 0.762 & 0.050 & 17.29 \\
$+$ SPADE\,\&\,RCASM prior         & 32.37 & 0.768 & 0.052 & 18.05 \\
\;$\hookrightarrow$ realism, always on      & 32.13 & 0.768 & 0.047 & 15.07 \\
\;$\hookrightarrow$ realism, gated in $t$   & 31.98 & 0.768 & 0.047 & 15.67 \\
\;$\hookrightarrow$ realism, gated in epoch & 32.16 & 0.773 & 0.045 & 15.21 \\
\midrule
\multicolumn{5}{l}{\footnotesize\emph{Shared model, both priors on, cumulative}} \\
Shared trunk, joint AdaGN          & 32.08 & 0.760 & 0.051 & 18.85 \\
$+$ separate AdaGN                 & 32.23 & 0.765 & 0.049 & 17.04 \\
$+$ realism, always on             & 32.03 & 0.764 & 0.052 & 19.34 \\
$+$ realism, gated in $t$           & 31.82 & 0.764 & 0.046 & 15.72 \\
\textbf{$+$ realism, gated in epoch} & 32.10 & 0.769 & 0.044 & 14.51 \\
\bottomrule
\end{tabular}
\end{minipage}
\caption{\textbf{Ablation, read separately in each direction.} 1181 held-out pairs, 100 epochs per direction. Upper block: one model per direction, the three $\hookrightarrow$ rows being alternative realism schedules on top of the physical prior. Lower block: one model serving both directions with both priors routed by direction, rows cumulative, the last being DARB, the same model in both panels. The schedules differ in weight as well as in timing, so the comparison is between whole schedules rather than single knobs.}
\label{tab:ablation}
\end{table*}

\begin{figure*}[t]
\centering
\begin{minipage}[t]{0.320\textwidth}\vspace{0pt}\centering
  \includegraphics[width=\linewidth]{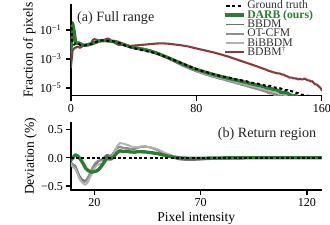}
  \caption{\textbf{Intensity statistics of generated sonar}, pooled over the held-out set. BDBM ($\dagger$) is omitted from (b), where its deviation runs off scale.}
  \label{fig:hist}
\end{minipage}\hfill
\begin{minipage}[t]{0.320\textwidth}\vspace{0pt}\centering
  \includegraphics[width=\linewidth]{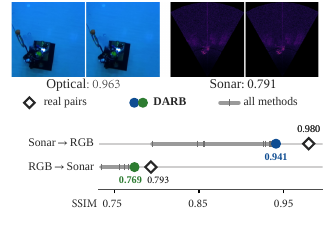}
  \caption{\textbf{What SSIM can resolve in each direction.} Top: one matched pair. Bottom: the mean over $400$ pairs; the grey span covers all of Tab.~\ref{tab:main}.}
  \label{fig:ssim}
\end{minipage}\hfill
\begin{minipage}[t]{0.320\textwidth}\vspace{0pt}\centering
  \resizebox{\linewidth}{!}{%
  \begin{tabular}{l cccc}
  \toprule
   & PSNR $\uparrow$ & SSIM $\uparrow$ & LPIPS $\downarrow$ & FID $\downarrow$ \\
  \midrule
  \multicolumn{5}{l}{\emph{Sonar $\rightarrow$ RGB}} \\
  BiBBDM     & 12.93 & 0.557 & 0.678 & 270.5 \\
  \;$+10$ ep & 13.97 & 0.542 & 0.621 & 260.3 \\
  DARB       & 13.68 & 0.576 & 0.649 & 270.4 \\
  \;\textbf{$+10$ ep} & \textbf{15.87} & \textbf{0.645} & \textbf{0.521} & \textbf{236.6} \\
  \midrule
  \multicolumn{5}{l}{\emph{RGB $\rightarrow$ Sonar}} \\
  BiBBDM     & 14.88 & 0.391 & 0.557 & 171.2 \\
  \;$+10$ ep & 15.75 & 0.387 & 0.510 & 161.6 \\
  DARB       & 15.30 & 0.428 & 0.514 & 191.2 \\
  \;\textbf{$+10$ ep} & \textbf{18.48} & \textbf{0.515} & \textbf{0.309} & \textbf{148.9} \\
  \bottomrule
  \end{tabular}}
  \captionof{table}{\textbf{Cross-Dataset Transfer}, 308 held-out pairs differing in field of view, water body and target class. Plain rows are zero-shot, indented rows add ten epochs per direction.}
  \label{tab:transfer}
\end{minipage}
\vspace{-0.2cm}
\end{figure*}

\paragraph{Effect of the direction path.}
Replacing the joint AdaGN of the shared trunk with a separate direction branch costs $12.9$\,M parameters, $4.3\%$ of the model, and gains $0.66$\,dB in the sonar-to-optical direction and $0.15$\,dB in the reverse, with LPIPS improving on both sides and FID moving by $-1.81$ in the reverse direction and $+0.13$ in the forward one (Tab.~\ref{tab:ablation}). Carried inside the timestep embedding, a binary direction label competes with a continuous high-variance signal; given its own branch it does not.

\paragraph{Effect of the physical priors.}
On a dedicated model the UAM prior gains $1.84$\,dB and $0.010$ SSIM in the sonar-to-optical direction at a cost of $4.43$ FID, and the acoustic pair gains $0.22$\,dB and $0.006$ SSIM in the reverse at a cost of $0.76$ FID. Two operators of different form produce the same trade, because a prior fixes what the scene should look like and in doing so suppresses the high-frequency content that FID rewards. Recovering it is what the realism term is for.



\paragraph{The realism schedule.}
Confining the perceptual term to a fixed diffusion-time window, $t \le T/4$, degrades every cell of the dedicated models, by $0.47$\,dB and $0.57$ FID forward and $0.15$\,dB and $0.60$ FID in reverse, yet gains $2.24$\,dB and $2.42$ FID forward and $3.62$ FID in reverse on the shared trunk. The restriction therefore treats divided capacity rather than a defect of the objective, and it is the schedule ARS replaces. Delaying the term instead of opening it at the first epoch improves all eight cells of the shared model, by up to $3.15$\,dB and $4.83$ FID, and unlike the fixed window it helps the dedicated models too, so what remains is which of two onsets each direction receives. ARS reads that from the force ratio of Eq.~\eqref{eq:sratio}: over the first $E_{\mathrm{win}}$ epoch-equivalents the median is $1.18$ for RGB to sonar and $0.44$ for sonar to RGB, so the first direction admits the term as soon as the window closes while the second is deferred to $E_{\mathrm{def}}$. Which direction waits is decided by that measurement, not by us; how long it waits is a fixed budget.

\subsection{Metric Behaviour and Cross-Dataset Transfer}
\label{sec:exp_instrument}

\paragraph{Why is SSIM asymmetric.}
SSIM separates the two directions of DARB by $0.172$, $0.941$ against $0.769$, where PSNR separates them by $0.19$\,dB. What differs is the target, not the quality of the prediction. Image the same patch twice: the objects and their shadows come back, the speckle does not. Call the part that returns $m$ and the rest $u$, so two captures are $x = m+u$ and $y = m+v$ with $u$ and $v$ independent. SSIM's cross-covariance then keeps only $\sigma_m^{2}$ while both variances keep $\sigma_m^{2}+\sigma_u^{2}$:
\begin{equation}
\mathrm{SSIM}_{\mathrm{ref}} = \frac{\sigma_m^{2}}{\sigma_m^{2} + \sigma_u^{2}} = \frac{1}{1+r}, \qquad r = \sigma_u^{2} / \sigma_m^{2}.
\label{eq:ssim_ceiling}
\end{equation}
Nothing can recover $u$, so this is what a perfect estimator earns. Speckle makes $u$ large in sonar; sensor noise makes it small in optics. The two directions therefore sit under different ceilings; we measure both. Among $10{,}629$ captures we pair frames by a $32\times32$ descriptor, coarse enough to average speckle away, and score each pair at 256×256, the resolution of Tab.~\ref{tab:main}; pairing on one modality and scoring the other keeps each number honest. Over $400$ pairs, two real captures of one patch score $0.980$ optically and $0.793$ acoustically (Fig.~\ref{fig:ssim}). The ceilings differ by $0.186$ and our two directions by $0.172$; every method in Tab.~\ref{tab:main} sits below both. The columns measure distance to different ceilings and cannot be averaged.

\paragraph{Cross-Dataset Transfer.}
Redeploying an underwater platform changes the field of view, the water body and the target class at once, and Tab.~\ref{tab:transfer} evaluates both unified models on a collection that differs in all three. Without adaptation DARB leads on every per-image metric in both directions and trails by $20.0$ FID in the acoustic one, so per-image and distributional agreement do not transfer together. Zero-shot is not the operating point, however: a platform entering new water is calibrated once, on whatever data the first survey returns, so the question is what a small fixed budget buys. Ten epochs per direction answer it differently for the two models. In SSIM, fine-tuning moves BiBBDM down in both directions, by $0.015$ and $0.004$, and moves DARB up in both, by $0.069$ and $0.087$; the two models improve not at different rates but in different directions, on the same data under the same protocol. The acoustic FID deficit reverses over the same epochs, from $20.0$ behind to $12.7$ ahead. Routing the physical priors therefore leaves a model that recalibrates within a commissioning budget rather than relearns the domain.

%% file: sec/5_Conclusion.tex
\section{Conclusion}
\label{sec:conclusion}
We address bidirectional sonar and optical translation with a single network, on the principle that unified deployment does not require identical treatment. DARB shares one diffusion-bridge trunk and introduces asymmetry only where the physics forces it: spatially, each direction is routed through the operator its formation model calls for; temporally, Adaptive Realism Supervision reads when the perceptual objective starts, which samples it acts on and how strongly, from the training loop itself. One network matches the sonar-to-optical specialist to within $0.11$\,dB PSNR, lowers optical-to-sonar FID by $0.70$, and improves on two independently trained bridges in seven of eight metrics at $37\%$ fewer parameters. Neither the onset nor the weight was set by hand; both are what one rule returns on each side. The bridge needs co-registered pairs, which deployments rarely provide; extending measured rather than designed asymmetry to unregistered settings is the natural next step.